\documentclass[letterpaper, 10 pt, conference,journal]{ieeeconf}  % Comment this line out if you need a4paper

\IEEEoverridecommandlockouts                              % This command is only needed if 
\usepackage{graphicx}
\usepackage{cite}
\usepackage{amsmath,amssymb}  % 处理数学公式
\usepackage{hyperref}  % 处理超链接（可选）
\usepackage{url}  % 处理 URL 链接
\usepackage{algorithm}  % 处理算法（如有）
\usepackage{algpseudocode} 
\usepackage{booktabs}
\usepackage{array}
\usepackage{adjustbox}
\usepackage{bm}
\usepackage{lmodern}
\title{\LARGE \bf
HTMNet: A Hybrid Network with Transformer-Mamba Bottleneck Multimodal Fusion for Transparent and Reflective Objects Depth Completion
}

\author{Guanghu Xie, Yonglong Zhang , Zhiduo Jiang, Yang Liu\textsuperscript{\dag}, Zongwu Xie, Baoshi Cao, Hong Liu
\thanks{\textsuperscript{\dag} Corresponding author:Yang Liu(liuyanghit@hit.edu.cn).}
\thanks{*This work was supported by the Natural Science Foundation of Heilongjiang Province for Excellent Young Scholars (Grant No. YQ2024E018) and the Youth Talent Support Program of the China (Grant No. 2022-JCJQ-QT-061).}
\thanks{All authors are with with the State Key Laboratory of Robotics and Systems, Harbin Institute of Technology, Harbin 150001, Heilongjiang, China
}
}

\begin{document}

\maketitle
\thispagestyle{empty}
\pagestyle{empty}

%%%%%%%%%%%%%%%%%%%%%%%%%%%%%%%%%%%%%%%%%%%%%%%%%%%%%%%%%%%%%%%%%%%%%%%%%%%%%%%%
\begin{abstract}
Transparent and reflective objects pose significant challenges 
for depth sensors, resulting in incomplete depth information 
that adversely affects downstream robotic perception 
and manipulation tasks. 
To address this issue, we propose HTMNet, 
a novel hybrid model integrating Transformer, CNN, 
and Mamba architectures.  
The encoder is based on a dual-branch CNN-Transformer framework, the bottleneck fusion module adopts a Transformer-Mamba architecture, and the decoder is built upon a multi-scale fusion module.
We introduce a novel multimodal fusion module grounded 
in self-attention mechanisms and state space models, 
marking the first application of the Mamba architecture 
in the field of transparent object depth completion 
and revealing its promising potential. 
Additionally, we design an innovative multi-scale fusion module 
for the decoder that combines channel attention, spatial attention, 
and multi-scale feature extraction techniques 
to effectively integrate multi-scale features 
through a down-fusion strategy. 
Extensive evaluations on multiple public datasets demonstrate 
that our model achieves state-of-the-art(SOTA) performance, 
validating the effectiveness of our approach. 
\end{abstract}


%%%%%%%%%%%%%%%%%%%%%%%%%%%%%%%%%%%%%%%%%%%%%%%%%%%%%%%%%%%%%%%%%%%%%%%%%%%%%%%%
% \section{INTRODUCTION}

\begin{figure}[htbp] % 使用figure*来跨双栏显示图片
    \centering
    % \begin{textblock*}{5cm}(0.8\textwidth,0.05\textheight)
    \includegraphics[width=\columnwidth]{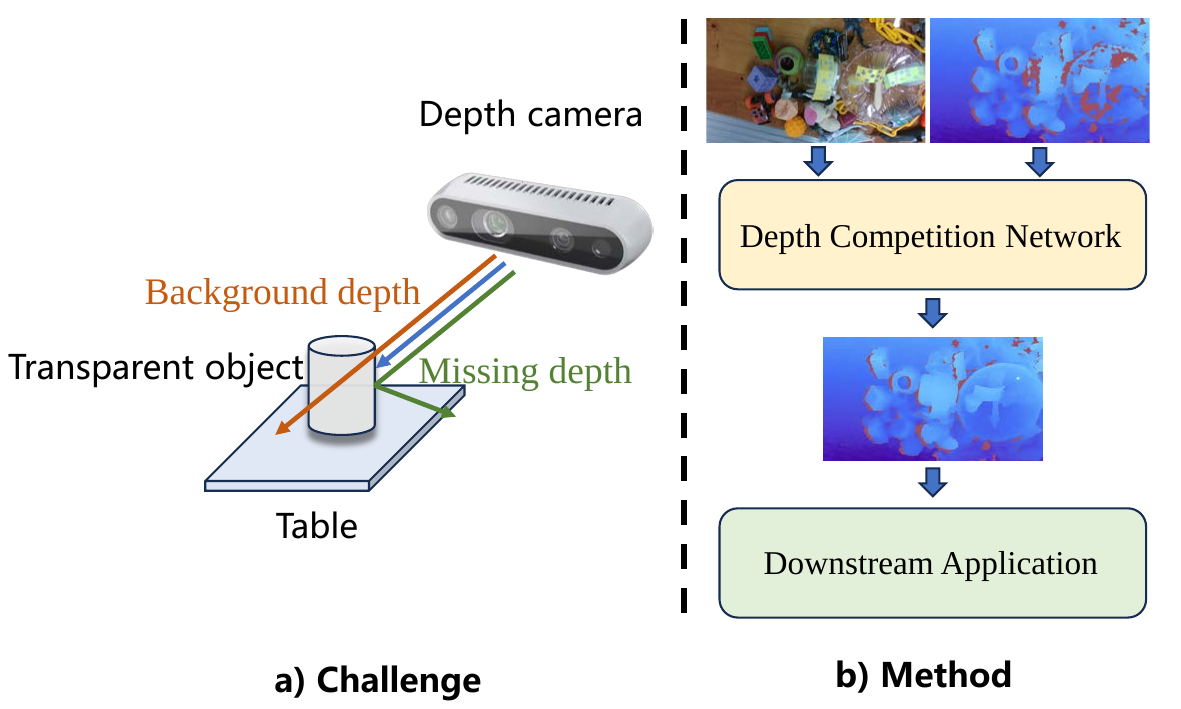} % 图片路径和宽度自行调整
    \caption{a$)$ illustrates two typical types of errors 
    encountered when capturing the depth 
    of transparent objects using depth cameras: 
    one is missing depth information, 
    and the other is the erroneous acquisition 
    of background depth;b$)$ depicts the general pipeline in which a depth completion model is used to recover the depth of transparent objects, which is subsequently fed into downstream tasks.}
    \label{fig0}
%    \end{textblock*}
\end{figure}

\section{INTRODUCTION}

The acquisition of accurate depth information for transparent 
and reflective objects remains a significant challenge 
in the field of computer vision\cite{cui2023light}. 
These objects exhibit unique optical characteristics 
due to their transparency and reflectivity, 
which conflict 
with common depth imaging assumptions\cite{sun2024diffusion}\cite{wang2025transdiff}. 
This conflict renders traditional depth imaging methods ineffective for such objects, 
making it difficult to obtain their 3D information\cite{liu2025gaa}. 
% Therefore, this study aims to explore a novel depth reconstruction method for transparent and reflective objects, aiming to overcome existing technological limitations and provide more precise solutions for relevant applications.

3D geometric data, such as point clouds and depth maps, 
play a crucial role in vision-based perception and detection tasks. 
Existing depth sensors, including LiDAR and depth cameras, 
primarily rely on infrared light to acquire geometric information. 
However, infrared light penetrates transparent objects 
and undergoes specular reflection on reflective surfaces, 
making depth estimation challenging and introducing substantial noise\cite{sun2024diffusion}\cite{dai2022domain}.
As illustrated in Fig.\ref{fig0} a), 
the introduction of background depth 
and missing depth are two typical challenges 
that lead to inaccurate depth measurements for transparent and reflective objects. 
% Depth completion aims to address this issue by learning 
% to recover the missing depth values of such objects, 
% thereby enabling accurate perception for downstream tasks, 
% as shown in Fig.\ref{fig0} b).

The causes of depth loss are multifaceted, 
involving both optical and geometric factors. 
Transparent and reflective objects attenuate light signals, 
while object occlusion and sensor disparity 
lead to depth information loss\cite{huang2024distillgrasp}. 
Even after calibration, misalignment between RGB and depth cameras results 
in optical parallax and missing depth map blocks\cite{costanzino2023learning}\cite{jin2017robust}. 

The difficulties in recovering the 3D information 
of transparent and reflective objects mainly stem from two aspects. 
First, the complex optical properties of these objects pose challenges 
for cameras when acquiring RGBD images\cite{wang2025transdiff}\cite{zhou2024transparent}\cite{yan2024transparent}. 
Transparent objects suffer from feature masking due to background color transmission 
and refractive distortion, while reflective objects exhibit issues 
such as specular highlights 
and environment mapping\cite{liu2025gaa}-\cite{zhou2024transparent}. 
These issues disrupt traditional depth information acquisition methods like structured light 
and stereo vision. Additionally, multipath effects 
and refractive errors introduce noise and artifacts, 
leading to edge blurring and topological errors in depth reconstruction.

\begin{figure}[htbp] % 使用figure*来跨双栏显示图片
    \centering
    % \begin{textblock*}{5cm}(0.8\textwidth,0.05\textheight)
    \includegraphics[width=\columnwidth]{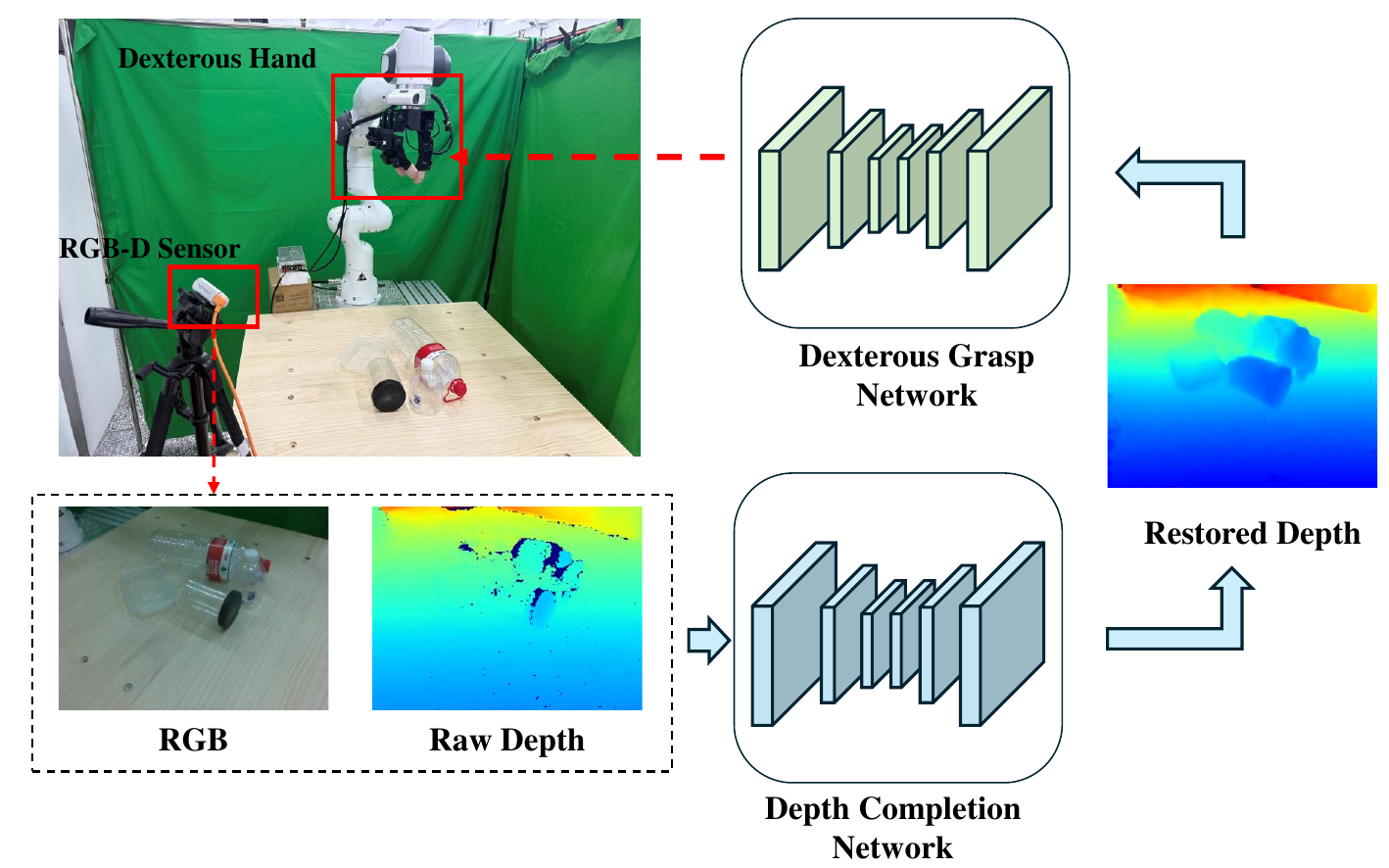} % 图片路径和宽度自行调整
    \caption{Depth completion plays a crucial role in dexterous grasping applications. When grasping transparent or specular objects, depth completion is first performed to obtain relatively complete depth information. The completed depth maps are then fed into a dexterous grasping network for grasp detection. This process effectively addresses the failure of grasp detection caused by missing depth information in transparent or specular objects.}
    \label{fig0_grasp}
%    \end{textblock*}
\end{figure}

% Depth completion aims to address these issues by learning 
% to recover the missing depth values of such objects, 
% thereby enabling accurate perception for downstream tasks, 
% as shown in Fig.\ref{fig0} b).
Numerous research efforts have been dedicated to acquiring depth information for transparent and reflective objects, including methodologies such as monocular depth estimation and depth completion based on incomplete depth data.
Monocular RGB image depth estimation 
is inherently ill-posed due to the lack 
of geometric constraints\cite{li2024segment}\cite{shen2024gamba}. 
To address this, sparse depth priors are introduced, 
transforming the task into a depth inpainting process 
from sparse to dense\cite{wang2025transdiff}. 
This approach leverages physically meaningful depth signals 
to guide the model, generating dense depth maps that conform 
to the real scene structure. This method enhances the stability 
and accuracy of depth reconstruction by integrating multimodal information.
Depth completion aims to recover the missing depth values of such objects through a learning-based approach, 
thereby enabling accurate perception for downstream tasks, 
as shown in Fig.\ref{fig0} b).
Fig.\ref{fig0_grasp} presents an application example of depth completion, where the completed depth of transparent and specular objects is used for multi-finger dexterous grasp detection. This facilitates a higher grasp success rate.

% The causes of depth loss are multifaceted, 
% involving both optical and geometric factors. 
% Transparent and reflective objects attenuate light signals, 
% while object occlusion and sensor disparity 
% lead to depth information loss\cite{huang2024distillgrasp}. 
% Even after calibration, misalignment between RGB and depth cameras results 
% in optical parallax and missing depth map blocks\cite{costanzino2023learning}\cite{jin2017robust}. 

Models based on Transformer and CNN have been widely applied in the field of transparent and reflective objects depth completion. To the best of our knowledge, however, there are currently no reported applications of the Mamba architecture in this domain. To address this long-standing gap, we conduct our research to explore the potential of the Mamba architecture in this context, aiming to open new research avenues for the community.

The main contributions of our work are summarized as follows:
\begin{itemize}
\item \textbf{We propose a novel dual-branch hybrid architecture integrating Transformer, CNN, and Mamba.} In this design, the Transformer branch extracts features from RGB-D images, the CNN branch extracts features from depth maps, and a bottleneck multimodal fusion is performed based on attention mechanisms and state-space models. This represents a new hybrid architecture tailored for depth completion of transparent objects.
        
\item \textbf{We develop a Transformer-Mamba-based bottleneck multimodal fusion module.} By leveraging self-attention mechanisms and state-space modeling, the module effectively fuses multimodal features, fully exploiting the complementary advantages of different paradigms. To the best of our knowledge, this is the first application of state-space models to the field of transparent object depth completion.

\item \textbf{We propose a novel multi-scale fusion module}. It integrates spatial attention, channel attention, and multi-scale feature fusion mechanisms into a unified down-fusion structure for the decoder.

\item We conduct extensive evaluations on multiple public datasets for transparent and reflective objects depth completion, achieving \textbf{state-of-the-art(SOTA)} performance.
\end{itemize}

% \begin{figure}[htbp] % 使用figure*来跨双栏显示图片
%     \centering
%     % \begin{textblock*}{5cm}(0.8\textwidth,0.05\textheight)
%     \includegraphics[width=\columnwidth]{pic_grasp.pdf} % 图片路径和宽度自行调整
%     \caption{Depth completion plays a crucial role in dexterous grasping applications. When grasping transparent or specular objects, depth completion is first performed to obtain relatively complete depth information. The completed depth maps are then fed into a dexterous grasping network for grasp detection. This process effectively addresses the failure of grasp detection caused by missing depth information in transparent or specular objects.}
%     \label{fig0_grasp}
% %    \end{textblock*}
% \end{figure}

\section{Related Works}

\begin{figure*}[htbp] % 使用figure*来跨双栏显示图片
    \centering
    \includegraphics[width=\textwidth]{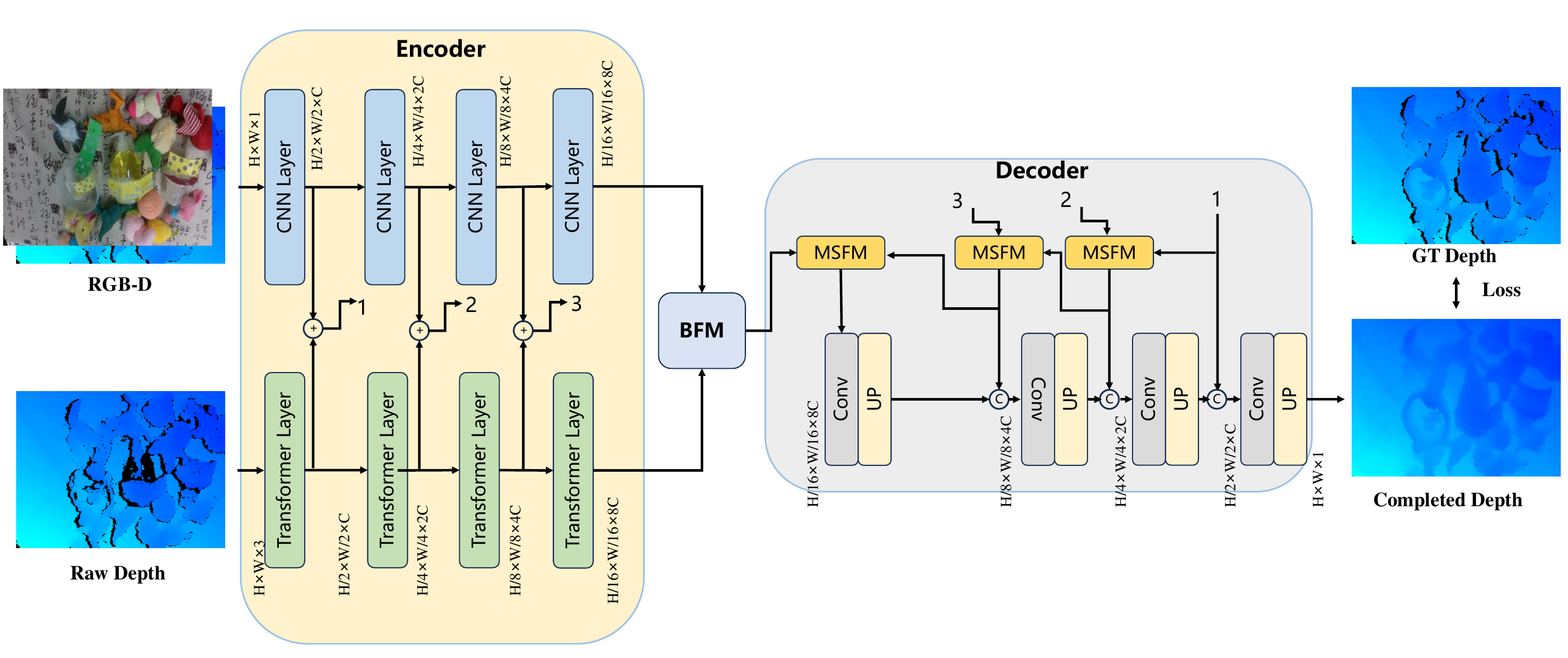} % 图片路径和宽度自行调整
    \caption{HTMNet Architecture.Our method consists of a dual-branch encoder, a bottleneck fusion module, and a decoder. The Transformer-based backbone extracts RGB-D features, while the CNN-based backbone extracts depth features. The bottleneck fusion module performs multimodal fusion at the network bottleneck, and the decoder is composed of a multi-scale fusion module, convolutional layers, and upsampling operations. 
    % b) Bottleneck fusion module(BFM): 
    % Composed of Transformer and Mamba modules, 
    % multimodal features are pixel-wise summed 
    % and then sequentially processed 
    % through a self-attention block, a Mamba block, 
    % and an MLP block to output enhanced fused features. 
    % c) Multi-scale fusion module(MSFM): 
    % Constructed based on spatial attention, 
    % channel attention, 
    % and multi-scale feature extraction mechanisms, 
    % it fuses multi-scale features from the encoder 
    % for the decoder.
    }
    \label{fig1}
\end{figure*}

\subsection{Single-View Depth Completion}
Depth completion, as a fundamental task in computer vision, 
typically uses raw depth and RGB images 
to recover missing depth values 
in depth maps\cite{wang2023decomposed}\cite{lin2023dyspn}\cite{chen2023agg}\cite{wang2022rgb}. 
Due to the convenience of single-view acquisition 
and rapid data capture, depth completion for transparent 
and reflective objects under single view settings 
has become a research hotspot. Before end-to-end models 
were applied to depth completion, two approaches 
were adopted for transparent object depth completion. 
The first relies on external assumptions, 
such as specialized sensing devices\cite{rantoson20103d} 
and predefined backgrounds\cite{wu2018full}. 
However, these methods are unsuitable for random environments 
due to their dependency on external assumptions. 
To overcome these limitations, 
a second approach 
using global optimization algorithms emerged\cite{zhang2018deep}. 
Studies in\cite{jiang2022a4t},\cite{sajjan2020clear}utilized 
sensor-captured RGB-D images as input for depth completion, 
eliminating the need for external assumptions. 
However, they rely on surface normal and edge predictions, 
a limitation later addressed by end-to-end models. 
In\cite{tang2021depthgrasp}, a generative adversarial network (GAN) 
was employed to generate depth maps 
for transparent object reconstruction. 
To enable cross-level feature fusion 
and better leverage low-level depth features, 
\cite{li2023fdct} introduces a skip-connected fusion branch 
within a U-Net framework. 
In\cite{fang2022transcg}, an end-to-end U-shaped depth completion framework is proposed, 
combining RGB and depth images as input. 
% In \cite{fan2024tdcnet}, a dual-branch CNN-Transformer architecture is being developed, accompanied by the design of a multi-scale fusion module that integrates features across different scales.
\cite{fan2024tdcnet} uses a dual-branch CNN-Transformer architecture and designs a multi-scale fusion module to integrate features at different scales.
% Although both ToDE-Trans\cite{chen2023tode} 
% and swinDRNet\cite{dai2022domain} adopt Swin Transformers, 
% they differ significantly in architecture. 
% Specifically, swinDRNet is a dual-stream fusion network 
% that integrates raw depth information 
% with predicted depth through confidence maps, 
% while ToDE-Trans follows a classical encoder-decoder structure. 
\cite{zhai2024tcrnet} designs a decoder 
with multiple cascaded refinement modules. 

% In\cite{li2023fdct}, to enable cross-level feature fusion 
% and better leverage low-level depth features, 
% researchers introduced a skip-connected fusion branch 
% within a U-Net framework. 

\subsection{Hybrid CNN-Transformer Architecture}
In visual recognition tasks, Vision Transformer (ViT) models 
have achieved remarkable success 
due to their superior performance\cite{dosovitskiy2020image}. 
However, the multi-head self-attention (MHA) mechanism 
not only incurs high computational costs 
but also struggles to capture inductive biases for local relationships. 
% To address this issue, numerous studies 
% have explored novel approaches 
% to integrate convolutional mechanisms with self-attention. 
Convolutional networks exhibit strong capabilities in capturing local features with relatively low computational cost. Consequently, integrating CNN and Transformer architectures has naturally emerged as a popular research direction, with numerous studies dedicated to leveraging the complementary strengths of both frameworks.
For instance, Li et al.\cite{li2023localvit}introduced MSA 
at the front end of the model while employing CNNs at the backend, 
a design that contrasts sharply 
with the method in\cite{peng2021conformer}, 
where CNNs first extract local features 
and MSA then learns long-range dependencies. 
The CNN-Transformer hybrid architecture 
has also garnered considerable attention 
in the fields of medical image processing and remote sensing. 
For example, TransUNet\cite{chen2021transunet} is the first work 
in medical image segmentation 
that integrates a U-shaped CNN with a Transformer, 
effectively combining local and global feature representations. 
In \cite{li2025cfformer}, novel channel-wise and spatial feature fusion modules are 
introduced to integrate local features extracted 
by the CNN branch and global context captured 
by the Transformer branch. 
Furthermore, \cite{ma2024multilevel} employs a dual-branch CNN 
to extract shallow multimodal features 
and leverages both self-attention 
and cross-attention mechanisms for deep multimodal fusion, 
achieving remarkable performance in the remote sensing domain.
% In certain specialized tasks, the parallel use of CNNs 
% and Transformers has gradually become a common strategy. 
% Taking\cite{chen2022mobile}as an example, 
% researchers typically extract features separately 
% through CNN and Transformer branches, followed by feature fusion 
% at later stages. 
% Additionally, in\cite{wang2021evolving}and\cite{yuan2021incorporating}, 
% convolutional and attention mechanisms are jointly utilized 
% to construct modules, which are iteratively applied 
% throughout the network. 
% Despite the diversity 
% of these fusion strategies, their core objective remains consistent: 
% enabling models to simultaneously learn global and local information.

\subsection{Vision Mamba}
Since its emergence, Mamba has garnered significant attention in the research community, with numerous efforts devoted to developing vision models based on the Mamba architecture\cite{xing2024segmamba}\cite{ma2024u}.
VMamba\cite{liu2024vmamba} is a vision-oriented model built upon the Mamba architecture, specifically designed for image processing tasks. It introduces a novel 2D Selective Scan mechanism that effectively expands the receptive field and aligns well with the structural characteristics and requirements of visual perception tasks.
MambaVision\cite{hatamizadeh2024mambavision} effectively integrates CNN and the Mamba architecture, enabling it to capture both long-range and short-range dependencies.
EfficientVMamba\cite{pei2025efficientvmamba} introduces a skip-sampling mechanism and a atrous-based selective scanning approach.
U-Mamba\cite{ma2024u} integrates state space models with convolutional neural networks (CNNs) and has been successfully applied to the field of medical image segmentation.
VM-UNet\cite{ruan2024vm} and Mamba-UNet\cite{wang2024mamba} construct complete encoder–decoder architectures based on state space models (SSMs).

\section{Method}
Our proposed model mainly consists of three components: a dual-branch Transformer-CNN encoder, a bottleneck fusion module based on the self-attention mechanism and state-space model, and a decoder based on a multi-scale fusion module. The following sections will provide a detailed introduction to each component.

% \begin{figure*}[htbp] % 使用figure*来跨双栏显示图片
%     \centering
%     \includegraphics[width=\textwidth]{pic_new_arch.pdf} % 图片路径和宽度自行调整
%     \caption{HTMNet Architecture.Our method consists of a dual-branch encoder, a bottleneck fusion module, and a decoder. The Transformer-based backbone extracts RGB-D features, while the CNN-based backbone extracts depth features. The bottleneck fusion module performs multimodal fusion at the network bottleneck, and the decoder is composed of a multi-scale fusion module, convolutional layers, and upsampling operations. 
%     % b) Bottleneck fusion module(BFM): 
%     % Composed of Transformer and Mamba modules, 
%     % multimodal features are pixel-wise summed 
%     % and then sequentially processed 
%     % through a self-attention block, a Mamba block, 
%     % and an MLP block to output enhanced fused features. 
%     % c) Multi-scale fusion module(MSFM): 
%     % Constructed based on spatial attention, 
%     % channel attention, 
%     % and multi-scale feature extraction mechanisms, 
%     % it fuses multi-scale features from the encoder 
%     % for the decoder.
%     }
%     \label{fig1}
% \end{figure*}

\begin{figure}[htbp] % 使用figure*来跨双栏显示图片
    \centering
    \includegraphics[width=\columnwidth]{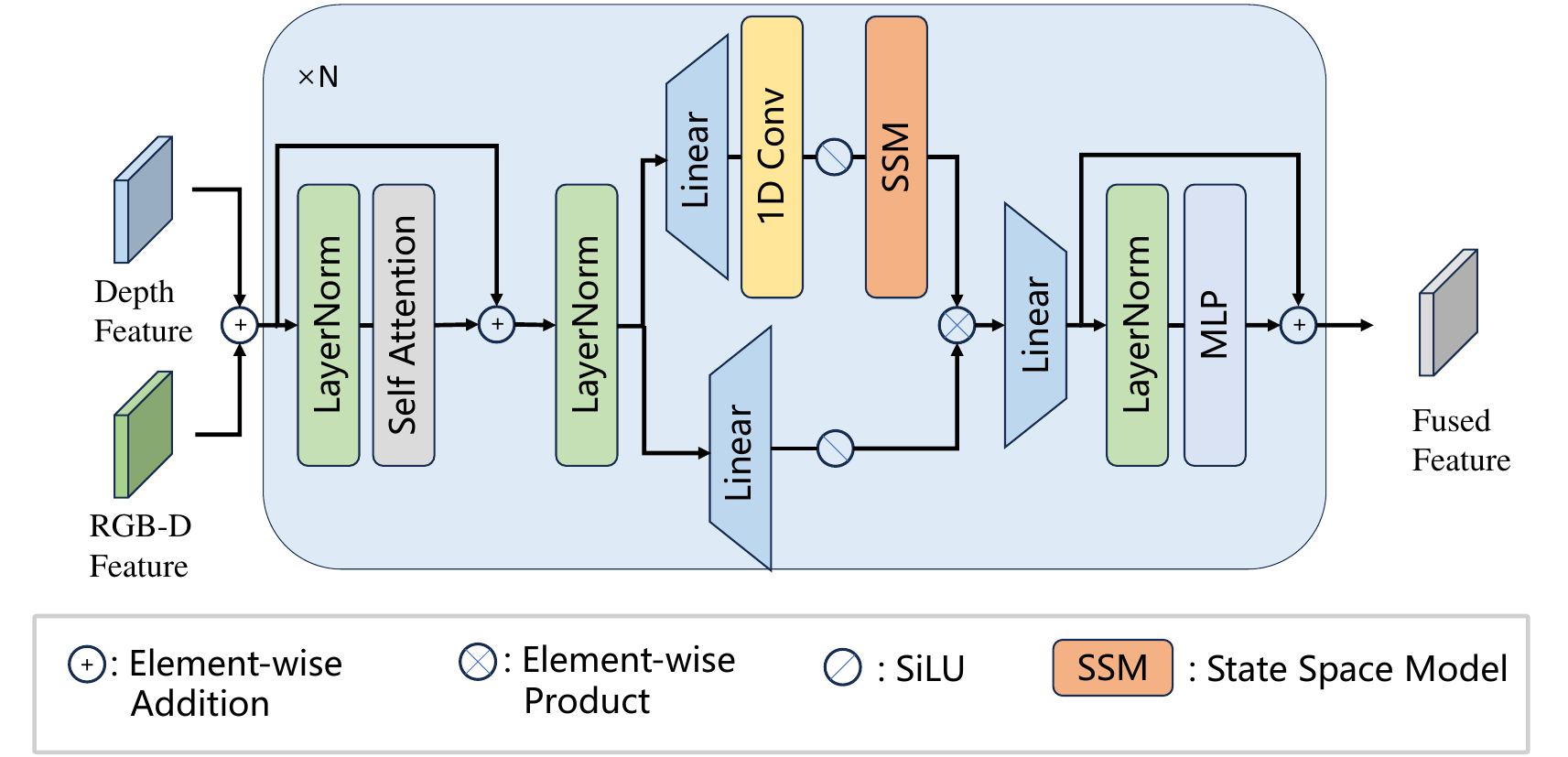} % 图片路径和宽度自行调整
    \caption{
        % HTMNet Architecture.Our method consists of a dual-branch encoder, a bottleneck fusion module, and a decoder. The Transformer-based backbone extracts RGB-D features, while the CNN-based backbone extracts depth features. The bottleneck fusion module performs multimodal fusion at the network bottleneck, and the decoder is composed of a multi-scale fusion module, convolutional layers, and upsampling operations. 
    Bottleneck fusion module(BFM): 
    Composed of Transformer and Mamba modules, 
    multimodal features are pixel-wise summed 
    and then sequentially processed 
    through a self-attention block, a Mamba block, 
    and an MLP block to output enhanced fused features. 
    % c) Multi-scale fusion module(MSFM): 
    % Constructed based on spatial attention, 
    % channel attention, 
    % and multi-scale feature extraction mechanisms, 
    % it fuses multi-scale features from the encoder 
    % for the decoder.
    }
    \label{fig1_bfm}
\end{figure}

\begin{figure}[htbp] % 使用figure*来跨双栏显示图片
    \centering
    \includegraphics[width=\columnwidth]{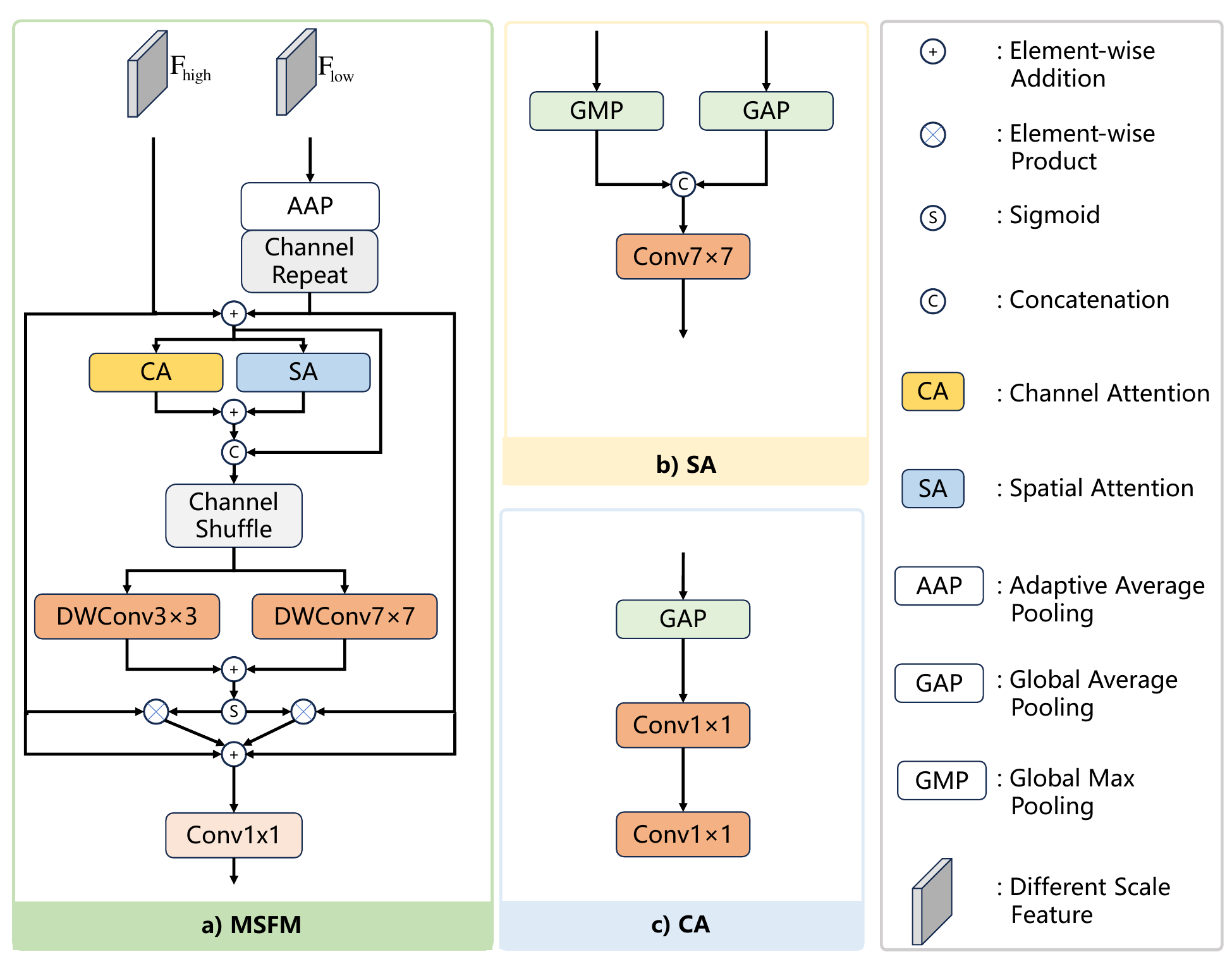} % 图片路径和宽度自行调整
    \caption{
        % HTMNet Architecture.Our method consists of a dual-branch encoder, a bottleneck fusion module, and a decoder. The Transformer-based backbone extracts RGB-D features, while the CNN-based backbone extracts depth features. The bottleneck fusion module performs multimodal fusion at the network bottleneck, and the decoder is composed of a multi-scale fusion module, convolutional layers, and upsampling operations. 
    % Bottleneck fusion module(BFM): 
    % Composed of Transformer and Mamba modules, 
    % multimodal features are pixel-wise summed 
    % and then sequentially processed 
    % through a self-attention block, a Mamba block, 
    % and an MLP block to output enhanced fused features. 
    Multi-scale fusion module(MSFM): 
    Constructed based on spatial attention, 
    channel attention, 
    and multi-scale feature extraction mechanisms, 
    it fuses multi-scale features from the encoder 
    for the decoder.
    }
    \label{fig1_msfm}
\end{figure}

\subsection{Dual-branch Transformer-CNN encoder}
Transformer tend to capture global features, 
while CNNs are more adept at capturing local features. 
To fully leverage the strengths of both, 
our proposed encoder consists of two branches: 
a Transformer branch and a CNN branch. 
The Transformer branch is utilized 
to extract RGB-D image features, 
while the CNN branch is employed to extract depth image features, 
similar to the approach in \cite{fan2024tdcnet}.

We adopt the Swin Transformer\cite{liu2021swin} as the backbone 
of the Transformer branch and ResNet\cite{he2016deep} as the backbone 
of the CNN branch to extract feature information 
from the two different modalities, respectively. 
Subsequently, we perform simple feature addition 
to fuse the shallow dual-modal features, 
while the deep features are fused 
using a bottleneck fusion module 
based on the self-attention mechanism and state-space model.

\subsection{Bottleneck fusion module}
The hybridization of Transformers and SSMs 
has been proven beneficial for long-sequence processing, 
and recent studies in the medical field \cite{yanhetero} 
have also demonstrated their potential in the visual domain. 
To explore their application potential 
in multimodal feature fusion and depth completion 
of transparent objects, inspired by \cite{yanhetero}, 
we designed a multimodal feature fusion module 
based on the self-attention mechanism and state-space model. 
This module is utilized at the bottleneck 
of the dual-branch backbone to fuse deep multimodal features. 
Fusion at the bottleneck, compared to layer-by-layer fusion, 
reduces computational costs while achieving higher performance, 
as validated in the supplementary experiments.
The following sections will provide a detailed introduction 
to this fusion module.

\begin{figure}[htbp] % 使用figure*来跨双栏显示图片
    \centering
    \includegraphics[width=\columnwidth]{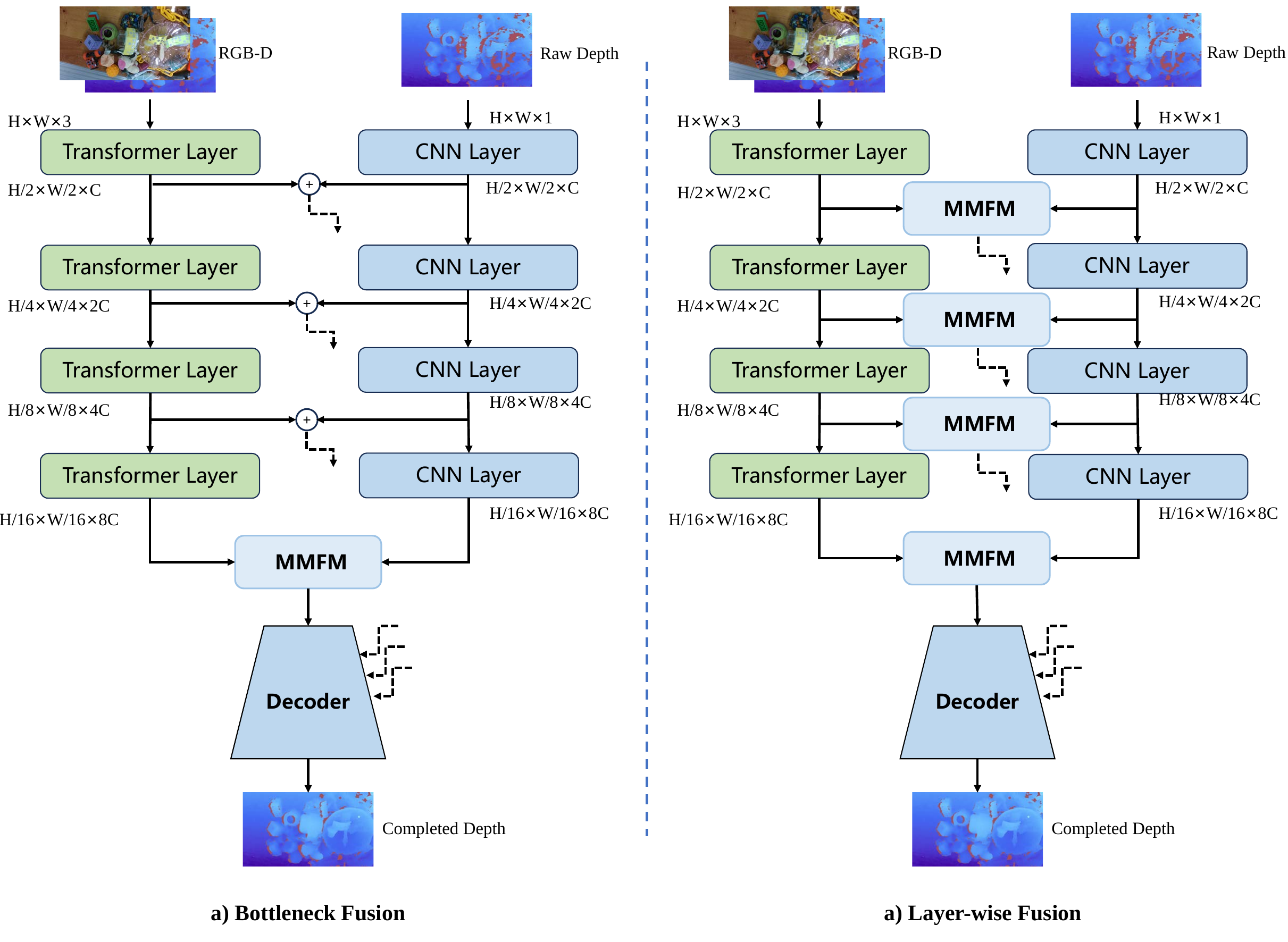} % 图片路径和宽度自行调整
    \caption{Comparison of Different Fusion Methods: 
    a) Fusion Module at Deep Bottleneck; 
    b) Layer-wise Fusion Modules. 
    MMFM Represents Multi-Modal Fusion Module.}
    \label{fig2}
\end{figure}

\textbf{Space-state Models(SSMs)}. 
We first introduce the preliminary knowledge of SSMs.
SSMs utilize a conventional continuous system 
that transforms a one-dimensional input function or sequence, represented as \(x(t) \in \mathbb{R}^N\), 
through intermediate implicit states \(h(t) \in \mathbb{R}^N\), 
to produce an output \(y(t) \in \mathbb{R}^N\).This process can be described by the following equation:

\begin{equation}
\label{equa0_0}
\begin{aligned}
     h'(t) &= \mathbf{A}h(t) + \mathbf{B}x(t)\\
      y(t) &= \mathbf{C}h(t)
\end{aligned}
\end{equation}

Where $\mathbf{A} \in \mathbb{R}^{N \times N}$ is the state transition matrix, 
while $\mathbf{B} \in \mathbb{R}^{N}$ and $\mathbf{C} \in \mathbb{R}^{N}$ are the input and output projection parameters, respectively.

Structured State Space Sequence Models(S4) and Mamba discretize this process 
to better suit deep learning applications.
An alternative approach is presented in the following equation:

\begin{equation}
\label{equa0_1}
\begin{aligned}
     \overline{\mathbf{A}} &= \text{exp}(\Delta\mathbf{A} ) \\
     \overline{\mathbf{B}} &= (\Delta\mathbf{A})^{-1}(\text{exp}(\Delta\mathbf{A})-I)\cdot \Delta\mathbf{B}\\
     h_t &= \overline{\mathbf{A}} h_{t-1} + \overline{\mathbf{B}} x_t\\
      y_t &= \mathbf{C}h_t
\end{aligned}
\end{equation}

Where $\Delta$ is the discretization interval. With the above preliminary knowledge, we next introduce the structure of our fusion module.

First, we perform element-wise addition of the RGB-D modality feature $X_\text{r} \in \mathbb{R}^{B \times C \times H \times W}$ 
and depth modality feature $X_\text{d} \in \mathbb{R}^{B \times C \times H \times W}$ to obtain the initial fused features:

\begin{equation}
\label{equa1}
    X_{\text{f}} = X_{\text{r}} + X_{\text{d}}
\end{equation}

Then, the initial fused features are passed 
through a multi-head self-attention mechanism(MHA) to extract global features:

\begin{equation}
\label{equa2}
    X_{\text{f}} = \text{LN}(X_{\text{f}}+\text{MHA}(X_{\text{f}}))
\end{equation}

Next, the fused feature $X_\text{f} \in \mathbb{R}^{B \times L \times C}$ is further enhanced through the Mamba block, 
which consists of two parallel branches. 
One branch expands the features to $(B, 2L, C)$ 
through a linear expansion layer, followed by a SiLU activation function. 
The other branch first expands the features through a linear layer, 
and then sequentially applies a 1D convolution, a SiLU activation function, 
and an SSM layer. 
Finally, the outputs of the two branches 
are fused using the Hadamard product:

\begin{equation}
\label{equa3}
    X_{\text{f}} = W_\text{down}(\text{SSM}(\delta(\text{Conv}(W_\text{up}X_{\text{f}})))) \odot  \delta(W_\text{up}X_{\text{f}})
\end{equation}

where \(\text{SSM}(\cdot)\) represents the state-space model. 
\(W_\text{up} \in \mathbb{R}^{C \times EC}\) 
and \(W_\text{down} \in \mathbb{R}^{EC \times C}\) are linear layers, 
where $E$ is expansion ratio. 
\(\text{Conv}(\cdot)\) denotes 1D convolution operation, 
\(\delta(\cdot)\) is SiLU activation function, 
and \(\odot\) represents element-wise multiplication.

Next, the fused features are fed into the MLP block:

\begin{equation}
\label{equa4}
    X_{\text{f}} = \text{LN}(X_{\text{f}}+\text{MLP}(X_{\text{f}}))
\end{equation}

\subsection{Multi-Scale Fusion Decoder}
Our decoder structure is inspired by \cite{fan2024tdcnet}. We integrate features of different scales through a multi-scale fusion module, as illustrated in Fig.\ref{fig1}. First, we perform downsampling and channel replication on the shallow features to ensure that the features from different scales have the same shape. Then, we compute the weights of features at different scales based on the spatial attention mechanism and the channel attention mechanism:

\begin{equation}
\label{equa5}
\begin{aligned}
    \hat{F}_{i-1} &= \text{CR}(\text{AAP}(F_{i-1}))\\
    X &= F_{i}+\hat{F}_{i-1} \\
    \text{SA}(X) &= \text{C}_{7\times 7}([\text{GMP}(X), \text{GAP}(X)]) \\
    \text{CA}(X) &= \text{C}_{1\times 1}(\text{C}_{1\times 1}(\text{GAP}(X)))
\end{aligned}
\end{equation}

where \(F_{i}\) denotes the feature map of the \(i\)-th layer and \(i \in \{1,2,3\}\). \(\text{CR}(\cdot)\) represents the channel repetition operation, and \(\text{AAP}(\cdot)\) denotes adaptive average pooling. \(\text{SA}(\cdot)\) and \(\text{CA}(\cdot)\) denote spatial attention and channel attention, respectively. \(\text{C}_{1\times 1}\) and \(\text{C}_{7\times 7}\) indicate convolution operations with kernel sizes of \(1\times 1\) and \(7\times 7\), respectively. \(\text{GAP}(\cdot)\) and \(\text{GMP}(\cdot)\) denote global average pooling and global max pooling, respectively.

To fully integrate the weights from different attention mechanisms, we perform channel shuffling to ensure thorough interleaving of the weights. Subsequently, the final weights are obtained through multi-scale DWConv and a Sigmoid activation function.

\begin{equation}
\label{equa6}
\begin{aligned}
   X_\text{w} &= \text{CS}([\text{SA}(X)+\text{CA}(X), X]) \\
   W &= \sigma (\text{DWConv}_{3\times 3}(X_w)+\text{DWConv}_{7\times 7}(X_w))\\
   X_\text{out} &= \text{C}_{1\times 1}(W\otimes F_{i})+\text{C}_{1\times 1}((1-W)\otimes \hat{F}_{i-1})
\end{aligned}
\end{equation}

Where $\otimes$ is element-wise product. 
$\text{DWConv}$ represents depth-wise convolution.
$\sigma$ is the sigmoid function.
$CS$ means channel shuffle operation.

\subsection{Loss function}
Similar to previous works \cite{fang2022transcg}\cite{fan2024tdcnet}, our loss function mainly consists of two parts. One is the MSE loss between the predicted depth values and the ground truth, which serves as the primary loss. Additionally, we employ a normal-based auxiliary smoothing loss function, as shown in the following equation:

\begin{equation}
\label{equa7}
\begin{aligned}
   L_{\text{mse}} &= \| D - D^{\ast }\|^{2} \\
   L_{\text{smooth}} &= 1-\text{cos}\langle V^{\ast }\times V\rangle  \\
   L_{\text{d}} &= L_{\text{mse}}+\alpha L_{\text{smooth}} \\
\end{aligned}
\end{equation}

where \(D\) and \(D^{\ast }\) denote the predicted and ground truth depth maps, respectively. \(V\) and \(V^{\ast }\) represent the predicted and ground truth normal vectors. \(\alpha\) is a balancing coefficient that controls the weight of the smoothing loss.

\begin{figure*}[htbp] % 使用figure*来跨双栏显示图片
    \centering
    \includegraphics[width=\textwidth]{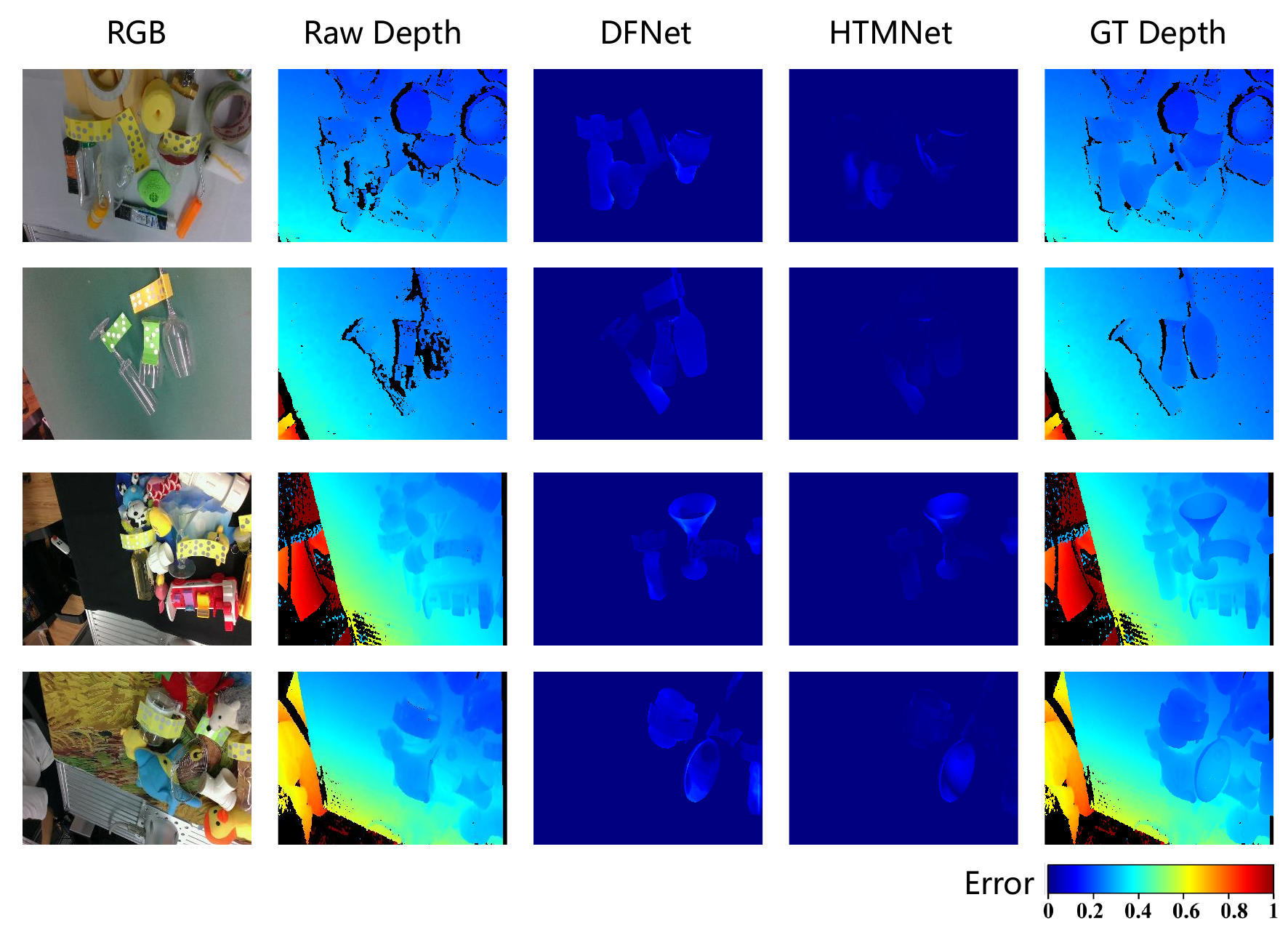} % 图片路径和宽度自行调整
    \caption{Depth Completion Visualizations of Different Models on the TransCG Dataset}
    \label{fig2_transcg}
\end{figure*}

\section{Experimental}
\subsection{Datasets and Metric}
\textbf{Datasets:}The datasets used for training and evaluating our model primarily include TransCG\cite{fang2022transcg}, ClearGrasp\cite{sajjan2020clear}, STD\cite{dai2022domain}.
The ClearGrasp dataset is synthesized using 9 real-world transparent plastic CAD models via the Synthesis AI platform, providing 23,524 samples for training. Its test set includes both synthetic and real-world transparent objects, covering seen and unseen shapes relative to the training set.
TransCG offers 57,715 RGB-D images collected by a robotic system in real-world settings. It focuses on 51 everyday objects prone to causing depth sensing errors, such as reflective, transparent, translucent objects, and items with dense holes.
STD dataset is a real-world dataset primarily consisting of specular, transparent, and diffuse objects. It is divided into two parts: known(STD-CatKnown) and novel categories(STD-CatNovel), and includes 30 scenes with a total of 27K RGB-D images.

\textbf{Metrics:}We evaluate our depth completion model using 4 widely recognized metrics: RMSE, REL, MAE, and threshold accuracy $\delta$. RMSE quantifies the root mean squared difference, while REL measures the mean absolute relative difference between predictions and ground-truth depths. MAE computes the mean absolute deviation. Threshold accuracy $\delta$ indicates the proportion of pixels satisfying $\max\left(\frac{d}{d^*}, \frac{d^*}{d}\right) < \delta$, where $d$ and $d^*$ represent the predicted and ground-truth depths, respectively. In our experiments, $\delta$ thresholds are set to 1.05, 1.10, and 1.25.

\subsection{Implementation Details}
The experiments were conducted on a system equipped with an Intel Xeon 8358P CPU and an Nvidia RTX 4090 GPU. We employed the AdamW optimizer with an initial learning rate of \(0.001\) to train the model for \(40\) epochs, using a batch size of \(8\). Input images were resized to \(320 \times 240\) pixels before being processed by the model.

\begin{figure*}[htbp] % 使用figure*来跨双栏显示图片
    \centering
    \includegraphics[width=\textwidth]{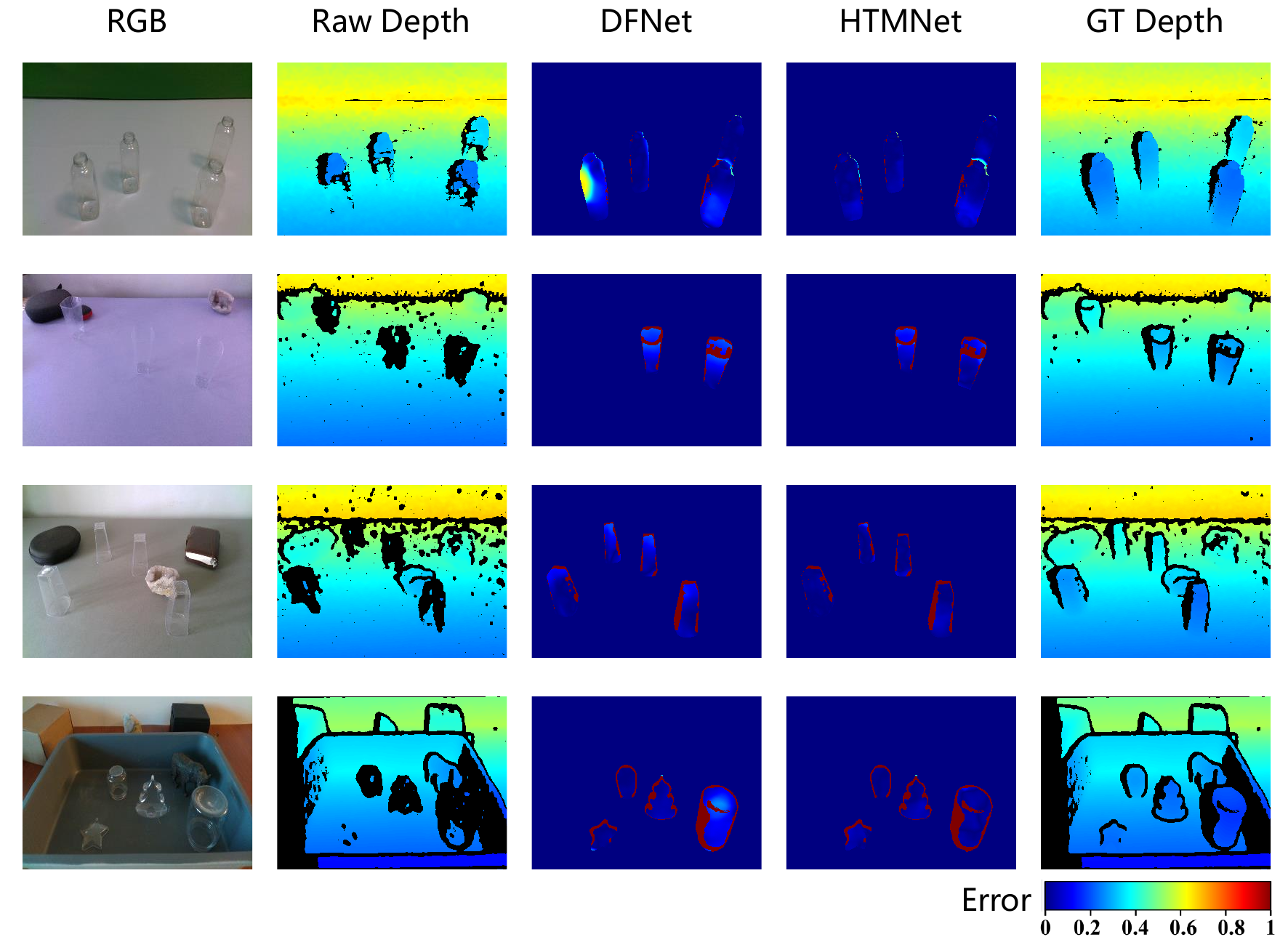} % 图片路径和宽度自行调整
    \caption{Depth Completion Visualizations of Different Models on the ClearGrasp Real-world Dataset}
    \label{fig_cg}
\end{figure*}

\subsection{Results and Analysis}
\subsubsection{TransCG Datasets}
We train our model on the TransCG training set 
and evaluate it on the corresponding test set. 
As shown in Tab.\ref{tab1}, 
our method achieves state-of-the-art performance on this dataset. 
Additionally, we visualize the prediction errors of different methods in Fig.\ref{fig2_transcg}, focusing exclusively on the transparent regions.
The error map is computed as $\frac{|d - d^*|}{d^*}$, where $d$ and $d^*$ denote the predicted depth and the ground truth, respectively.
It can be observed 
that our model produces the most detailed and accurate depth estimations, 
further demonstrating the effectiveness and superiority of our approach.

% \begin{figure*}[htbp] % 使用figure*来跨双栏显示图片
%     \centering
%     \includegraphics[width=\textwidth]{pic_transcg_vis_big.pdf} % 图片路径和宽度自行调整
%     \caption{Depth Completion Visualizations of Different Models on the TransCG Dataset}
%     \label{fig2_transcg}
% \end{figure*}

\begin{table}[htbp]
    \centering
    \caption{Performance comparison of different methods on TransCG dataset}
    \label{tab1}
    \resizebox{\columnwidth}{!}{
    \begin{tabular}{lcccccc}
    \toprule
    \textbf{Methods} & \textbf{RMSE} $\downarrow$ & \textbf{REL} $\downarrow$ & \textbf{MAE} $\downarrow$ & \textbf{$\delta_{1.05}$}$\uparrow$ & \textbf{$\delta_{1.10}$}$\uparrow$ & \textbf{$\delta_{1.25}$}$\uparrow$ \\
    \midrule
    CG\cite{sajjan2020clear}         & 0.054  & 0.083  & 0.037  & 50.48  & 68.68  & 95.28  \\
    DFNet\cite{fang2022transcg}      & 0.018  & 0.027  & 0.012  & 83.76  & 95.67  & 99.71  \\
    LIDF\cite{zhu2021rgb} & 0.019  & 0.034  & 0.015  & 78.22  & 94.26  & 99.80  \\
    TCRNet\cite{zhai2024tcrnet}     & 0.017  & 0.020  & 0.010  & 88.96  & 96.94  & $\mathbf{99.87}$  \\
    TranspareNet\cite{xu2021seeing} & 0.026  & 0.023  & 0.013  & 88.45  & 96.25  & 99.42  \\
    FDCT\cite{li2023fdct}       & 0.015  & 0.022  & 0.010  & 88.18  & 97.15  & 99.81  \\
    TODE-Trans\cite{chen2023tode}  & $0.013$  & 0.019  & $\mathbf{0.008}$  & 90.43  & 97.39  & 99.81  \\
    DualTransNet\cite{liu2024transparent} & 0.012  & 0.018  & $\mathbf{0.008}$  & $92.37$  & 97.98  & 99.81  \\
    TDCNet\cite{fan2024tdcnet} &$\mathbf{0.012}$ &$\mathbf{0.017}$ &$\mathbf{0.008}$ &92.25 &97.86 &99.84 \\
    \midrule
    HTMNet (ours) & $\mathbf{0.012}$ & $0.018$ & $\mathbf{0.008}$ & $\mathbf{92.40}$ & $\mathbf{98.17}$ & $\mathbf{99.87}$ \\
    \bottomrule
    \end{tabular}}
\end{table}

\subsubsection{Cleargrasp Datasets}
% Similar to previous works \cite{chen2023tode}\cite{fan2024tdcnet}, 
We train our model on ClearGrasp training sets and evaluate it 
on the ClearGrasp real-world test and val set. 
The experimental results are shown in Tab.\ref{tab2}. 
The CG\cite{sajjan2020clear} results are derived from the evaluation 
of its publicly available pre-trained model, 
while the results of DFNet\cite{fang2022transcg}, TDCNet\cite{fan2024tdcnet}, 
and FDCT\cite{li2023fdct} are obtained by retraining with the official open-source code and hyperparameters. 
% FDCT only provides the model code. Since its structure is similar to that of DFNet, we used DFNet's training code.
The DITR results are taken from the official paper.
% Our method outperforms most existing models, 
% achieving superior performance.
The experimental results demonstrate 
that our method surpasses many existing state-of-the-art approaches 
and achieves superior performance.
Furthermore, we visualize the prediction error in Fig.\ref{fig_cg}. 
The comparison of metrics and the visualization 
of depth completion results clearly demonstrate the superiority of our method.

% \begin{figure*}[htbp] % 使用figure*来跨双栏显示图片
%     \centering
%     \includegraphics[width=\textwidth]{pic_cg_vis_big.pdf} % 图片路径和宽度自行调整
%     \caption{Depth Completion Visualizations of Different Models on the ClearGrasp Real-world Dataset}
%     \label{fig_cg}
% \end{figure*}

\begin{table}[ht]
        \centering
        \caption{Comparison of Different Methods on Real-world ClearGrasp Dataset.}
        \resizebox{\columnwidth}{!}{
        \begin{tabular}{lcccccc}
        \toprule
        \textbf{Method} & \textbf{RMSE} & \textbf{REL} & \textbf{MAE} & $\boldsymbol{\delta_{1.05}}$ & $\boldsymbol{\delta_{1.10}}$ & $\boldsymbol{\delta_{1.25}}$ \\
        \midrule
        % DeepCompletion~\cite{zhang2018deep} & 0.209 & 0.396 & 0.207 & 34.61 & 52.79 & 71.32 \\
        % DenseDepth~\cite{alhashim2018high}  & 0.057 & 0.083 & 0.059 & 41.82 & 64.48 & 90.35 \\
        % SRD~\cite{song2021monocular}        & 0.049 & 0.072 & 0.044 & 67.11 & 79.64 & 91.33 \\
        % MiDaS~\cite{ranftl2020towards}      & 0.044 & 0.069 & 0.038 & 72.87 & 88.12 & 94.37 \\
        % LDM~\cite{rombach2022high}          & 0.046 & 0.071 & 0.044 & 74.18 & 83.57 & 92.19 \\
        CG~\cite{sajjan2020clear} & 0.037 & 0.049 & 0.027 & 74.30 & 88.47 & 96.27 \\
        
        % LIDF~\cite{zhu2021rgb}           & 0.028 & 0.035 & 0.022 & 79.17 & 91.14 & 98.30 \\
        % TranspareNet~\cite{xu2021seeing} & 0.026 & 0.039 & 0.022 & 76.93 & 90.02 & 98.10 \\
        DFNet~\cite{fang2022transcg}         & 0.026 & 0.037 & 0.020 & 76.69 & 92.26 & 99.09 \\
        FDCT\cite{li2023fdct}       & 0.028  & 0.038  & 0.021  & 76.45  & 93.36  & 98.95  \\
        TDCNet\cite{fan2024tdcnet} &$0.022$ &$0.031$ &$0.017$ &82.26 &95.83 &$\mathbf{99.85}$ \\
        DITR\cite{sun2024diffusion}               & $\mathbf{0.019}$ & 0.030 & $\mathbf{0.012}$ & 85.11 & 94.20 & 98.92 \\
        \midrule
        HTMNet(ours) & 0.020 & $\mathbf{0.026}$ & 0.015 & $\mathbf{86.32}$ & $\mathbf{96.32}$ & $99.67$ \\
        \bottomrule
        \end{tabular}}
        \label{tab2}
\end{table}

\subsubsection{STD Datasets}
\begin{figure*}[htbp] % 使用figure*来跨双栏显示图片
    \centering
    \includegraphics[width=\textwidth]{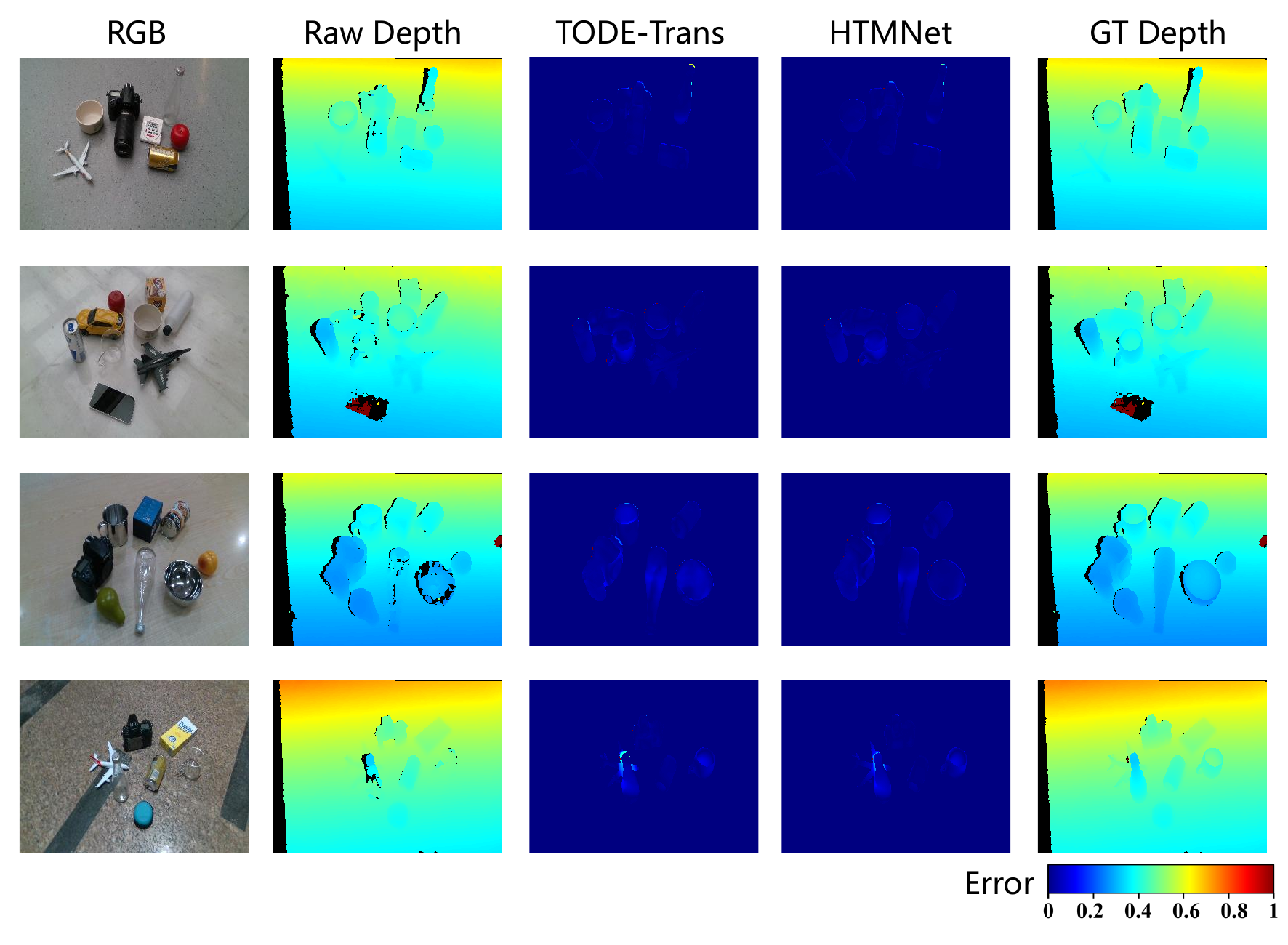} % 图片路径和宽度自行调整
    \caption{Depth Completion Visualizations of Different Models on the STDS Real-world Dataset}
    \label{fig_std}
\end{figure*}

We train our model on the STD-CatKnown training set and evaluate it on the test sets of both STD-CatKnown and STD-CatNovel. The experimental results are presented in the Tab.\ref{tab3_std}. For a fair comparison, we retrain other state-of-the-art methods listed in the table using the same training data. We also visualize the prediction results of different methods on the STD dataset, as shown in Fif.\ref{fig_std}. Both the quantitative results and qualitative visualizations demonstrate the superiority of our proposed method.

\begin{table}[ht]
        \centering
        \caption{Comparison of Different Methods on Real-world STD Dataset.}
        \resizebox{\columnwidth}{!}{
        \begin{tabular}{lcccccc}
        \toprule
        \textbf{Method} & \textbf{RMSE} & \textbf{REL} & \textbf{MAE} & $\boldsymbol{\delta_{1.05}}$ & $\boldsymbol{\delta_{1.10}}$ & $\boldsymbol{\delta_{1.25}}$ \\
        \midrule
        DFNet~\cite{fang2022transcg}         &0.025  &0.026  &0.016  &87.60  &95.92  &99.05  \\
        FDCT\cite{li2023fdct}       & 0.024  & 0.026  & 0.015  & 88.17  & 96.68  & 99.35  \\
        TODE-Trans\cite{chen2023tode}  & 0.021  & 0.022  & $\mathbf{0.013}$  & 90.31  & 97.39  & 99.69  \\
        \midrule
        HTMNet(ours) & $\mathbf{0.019}$ & $\mathbf{0.021}$ & $\mathbf{0.013}$ & $\mathbf{91.26}$ & $\mathbf{97.53}$ & $\mathbf{99.72}$ \\
        \bottomrule
        \end{tabular}}
        \label{tab3_std}
\end{table}

\subsection{Limits and Discussion}
\begin{figure*}[htbp] % 使用figure*来跨双栏显示图片
    \centering
    \includegraphics[width=\textwidth]{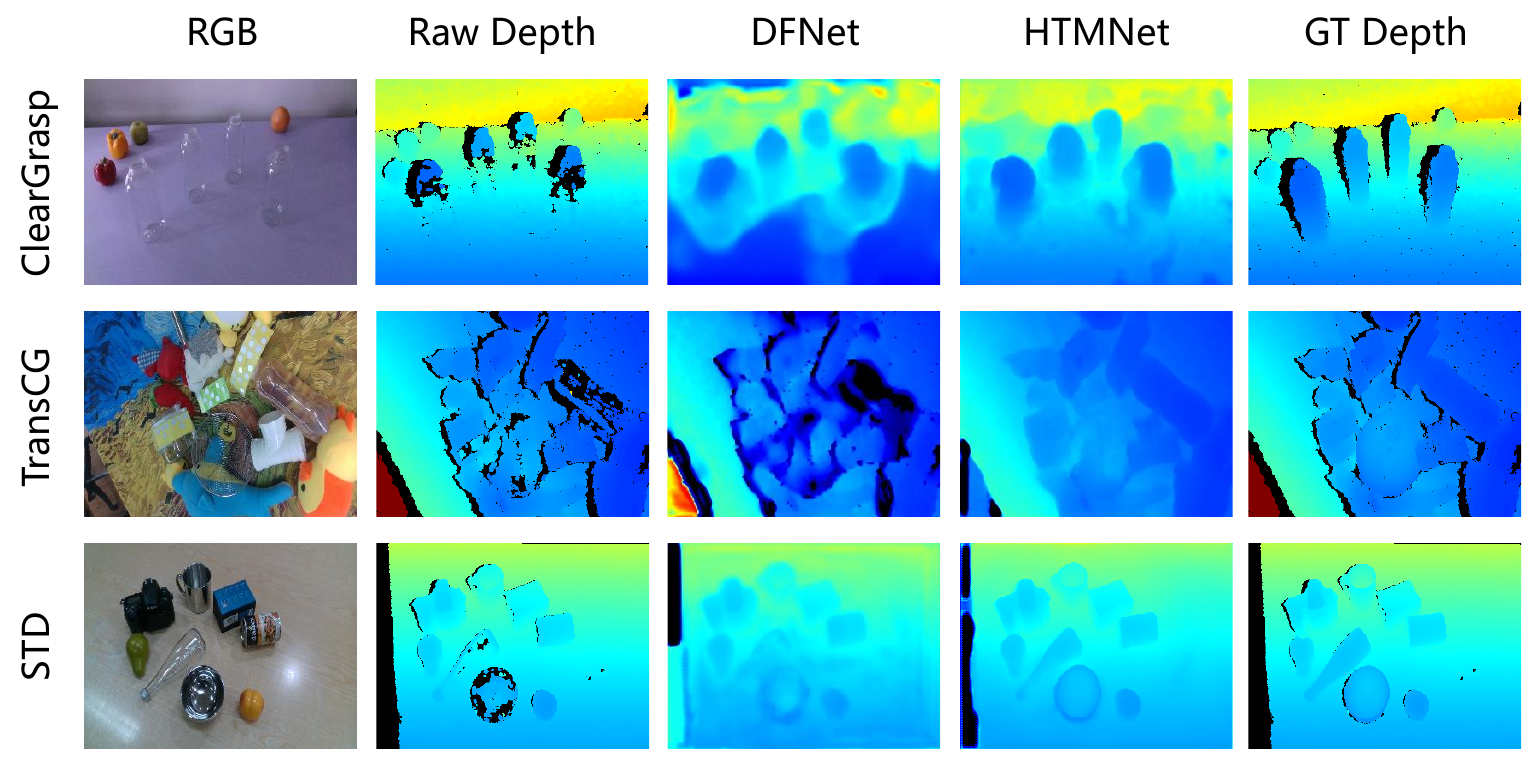} % 图片路径和宽度自行调整
    \caption{Depth Completion Visualizations on the Different Real-world Dataset}
    \label{fig_pred_gt}
\end{figure*}

Although our method achieves state-of-the-art 
performance in depth completion for transparent 
and reflective objects, 
it still exhibits certain limitations. 
Specifically, by prioritizing the completion accuracy 
of transparent and reflective regions through 
reduced training loss weights for non-transparent 
and non-reflective areas, there exists 
an inevitable degradation in depth precision 
for other regions. As illustrated in the Fig.\ref{fig_pred_gt}, 
while the original depth values 
of diffuse objects remain accurate, 
such compromised depth integrity 
during practical applications 
could adversely affect downstream tasks like robotic grasping 
by reducing success rates. 
This issue, however, remains a common limitation prevalent 
in existing methods. 
A potential solution involves incorporating diffuse objects 
into the loss computation paradigm, 
as exemplified by the training methodology implemented in the STD dataset,
which jointly optimizes depth estimation 
for three distinct categories: transparent, reflective, 
and diffuse objects.
As illustrated in Fig.\ref{fig_pred_gt}, the STD dataset is trained 
on objects spanning all categories present on the desktop surface. 
This approach ensures reasonable depth accuracy for opaque and non-reflective objects while effectively recovering missing depth information for transparent and reflective objects. Compared to model predictions derived from the TransCG and ClearGrasp datasets, the depth maps generated by the STD dataset exhibit significantly closer alignment with the ground-truth (GT) depth images.

\section{Ablation Study}
\subsection{Layer-wise Fusion vs. Bottleneck Fusion}
To evaluate the advantages of the proposed bottleneck fusion(BF) strategy, 
we design a layer-wise fusion(LWF) structure as a baseline, 
where the multimodal fusion module is applied at each stage 
of the encoder. As shown in Tab.\ref{tab3}, 
experimental results indicate 
that fusion at the bottleneck 
yields marginally superior overall performance 
compared to layer-wise fusion, 
along with improved inference throughput.

\begin{table}[htbp]
    \centering
    \caption{Performance comparison of different fusion motheds on TransCG dataset}
    \label{tab3}
    \resizebox{\columnwidth}{!}{
    \begin{tabular}{lcccccccc}
    \toprule
    \textbf{Model} & \textbf{RMSE} $\downarrow$ & \textbf{REL} $\downarrow$ & \textbf{MAE} $\downarrow$ & \textbf{$\delta_{1.05}$}$\uparrow$ & \textbf{$\delta_{1.10}$}$\uparrow$ & \textbf{$\delta_{1.25}$}$\uparrow$ &Param(M) &Throughput(Img/Sec) \\
    \midrule
    LWF & 0.012 & $0.018$ & $0.008$ & $\mathbf{92.51}$ & $97.94$ & $99.85$ &8.57 &230 \\
    BF & 0.012 & $0.018$ & $0.008$ & $92.40$ & $\mathbf{98.17}$ & $\mathbf{99.87}$ &11.31 &286\\
    \bottomrule
    \end{tabular}}
\end{table}

\subsection{Number of Bottleneck Fusion Modules}
To investigate the impact of the number 
of TransMamba modules at the bottleneck on model performance, 
we conducted a series of ablation studies. 
As shown in Tab.\ref{tab4},
experimental results show that the best performance 
is achieved when the number of modules is set to 4.

\begin{table}[htbp]
    \centering
    \caption{Performance Comparison of Different Number of Bottleneck Fusion Modules on TransCG Dataset}
    \label{tab4}
    \resizebox{\columnwidth}{!}{
    \begin{tabular}{lcccccc}
    \toprule
    \textbf{Num.} & \textbf{RMSE} $\downarrow$ & \textbf{REL} $\downarrow$ & \textbf{MAE} $\downarrow$ & \textbf{$\delta_{1.05}$}$\uparrow$ & \textbf{$\delta_{1.10}$}$\uparrow$ & \textbf{$\delta_{1.25}$}$\uparrow$ \\
    \midrule
    2 & $\mathbf{0.012}$ & $\mathbf{0.018}$ & $\mathbf{0.008}$ & $92.21$ & $97.85$ & $\mathbf{99.89}$ \\
    4 & $\mathbf{0.012}$ & $\mathbf{0.018}$ & $\mathbf{0.008}$ & $\mathbf{92.40}$ & $\mathbf{98.17}$ & $99.87$ \\
    6 & $\mathbf{0.012}$ & $\mathbf{0.018}$ & $\mathbf{0.008}$ & $92.34$ & $98.04$ & $\mathbf{99.89}$ \\
    \bottomrule
    \end{tabular}}
\end{table}

\subsection{The Role of Each Module}
We conducted controlled experiments on different modules, 
as shown in Tab.\ref{tab5}. It can be observed from the table 
that both the multi-scale fusion module 
and the bottleneck fusion module contribute 
to improving the depth completion performance of the model.

\begin{table}[htbp]
    \centering
    \caption{The Role of Different Modules on TransCG Dataset}
    \label{tab5}
    \resizebox{\columnwidth}{!}{
    \begin{tabular}{cccccccc}
    \toprule
    \textbf{MSFM} &\textbf{BFM} & \textbf{RMSE} $\downarrow$ & \textbf{REL} $\downarrow$ & \textbf{MAE} $\downarrow$ & \textbf{$\delta_{1.05}$}$\uparrow$ & \textbf{$\delta_{1.10}$}$\uparrow$ & \textbf{$\delta_{1.25}$}$\uparrow$ \\
    \midrule
     & & $0.013$ & $\mathbf{0.018}$ & $\mathbf{0.008}$ & $\mathbf{91.82}$ & $97.74$ & $99.86$ \\
    $\surd$  & & $\mathbf{0.012}$ & $\mathbf{0.018}$ & $\mathbf{0.008}$ & $92.35$ & $98.14$ & $\mathbf{99.89}$ \\
    $\surd$  &$\surd$ & $\mathbf{0.012}$ & $\mathbf{0.018}$ & $\mathbf{0.008}$ & $\mathbf{92.40}$ & $\mathbf{98.17}$ & $99.87$ \\
    \bottomrule
    \end{tabular}}
\end{table}

\section{Conclusions}
In this work, we propose a hybrid model named HTMNet for depth completion of transparent objects. HTMNet combines multiple architectures, including Transformer, CNN, and Mamba, to enhance model performance. We design a multimodal bottleneck fusion module based on self-attention and state space models, as well as a novel multi-scale fusion module specifically for multi-scale fusion in the decoder. We evaluate our method on several public datasets, achieving state-of-the-art performance. Both quantitative evaluations and visualization results demonstrate the effectiveness of our approach. In the future, we aim to further promote the application of transparent object depth completion in other domains, such as robotic grasping and pose estimation.

\bibliographystyle{IEEEtran}
\bibliography{refs}

\end{document}